\pdfoutput=1

\documentclass[11pt]{article}

\usepackage{emnlp2021}

\usepackage{times}
\usepackage{latexsym}

\usepackage[T1]{fontenc}

\usepackage[utf8]{inputenc}

\usepackage{microtype}

\usepackage{hyperref}
\usepackage{url}

\usepackage{multirow}
\usepackage{xcolor}
\usepackage{microtype}
\usepackage{graphicx}
\usepackage[normalem]{ulem}
\usepackage{amsmath}
\usepackage{amssymb}
\usepackage{multirow}
\usepackage{subcaption}

\usepackage[linesnumbered,ruled,vlined,english,onelanguage]{algorithm2e}

\title{Infusing Future Information into Monotonic Attention Through Language Models}

\author{Mohd Abbas Zaidi\thanks{$^{\star}$Equal contribution} , Sathish Indurthi\footnotemark[1] \thanks{$^{\dagger}$work was done at Samsung Research}, Beomseok Lee \\ {\bf Nikhil Kumar Lakumarapu}, {\bf Sangha Kim} \\
        NLP Lab, Samsung Research, Seoul, South Korea \\ \small{abbas.zaidi@samsung.com, sathish.indurthi@gmail.com, \{bsgunn.lee, n07.kumar, sangha01.kim1\} @samsung.com }}

\begin{document}
\maketitle
\begin{abstract}
Simultaneous neural machine translation (SNMT) models start emitting the target sequence before they have processed the source sequence. The recent adaptive policies for SNMT use monotonic attention to perform \textit{read/write} decisions based on the partial source and target sequences. The lack of sufficient information might cause the  monotonic attention to take poor \textit{read/write} decisions, which in turn negatively affects the performance of the SNMT model. On the other hand, human translators
make better \textit{read/write} decisions since they can anticipate the immediate future words using linguistic information and domain knowledge.
Motivated by human translators, in this work, we propose a framework to aid monotonic attention with an external language model to improve its decisions. We conduct experiments on the MuST-C English-German and English-French speech-to-text translation tasks to show the effectiveness of the proposed framework. The proposed SNMT  method  improves the quality-latency trade-off over the state-of-the-art monotonic multihead  attention.

\end{abstract}

\section{Introduction}
Simultaneous Neural Machine Translation (SNMT) addresses the problem of real-time interpretation in machine translation. A typical application of real-time translation is conversational speech or live video caption translation. In order to achieve live translation, an SNMT model alternates between reading from the source sequence and writing to the target sequence using either a fixed or an adaptive \textit{read/write} policy.

The fixed policies \cite{ma2019stacl} may introduce too much delay for some examples or not enough for others. The recent works focus on training adaptive policies using techniques such as monotonic attention. There are several variants of monotonic attention: hard monotonic attention \cite{raffel17a}, monotonic chunkwise attention \cite{chiu*2018monotonic}, monotonic infinite lookback attention (MILk) \cite{arivazhagan-etal-2019-monotonic}, and monotonic multihead attention (MMA) \cite{ma2019monotonic}. These monotonic attention processes can anticipate target words using only the available prefix source and target sequence. However, human translators anticipate the target words using their language expertise (linguistic anticipation) and contextual information (extra-linguistic anticipation) \cite{RAEI2001-n14-anticipation-in-conference-interpreting-a-cognitive-process} as well. 

\begin{figure}[t]
    \centering
    \includegraphics[width=0.53\textwidth]{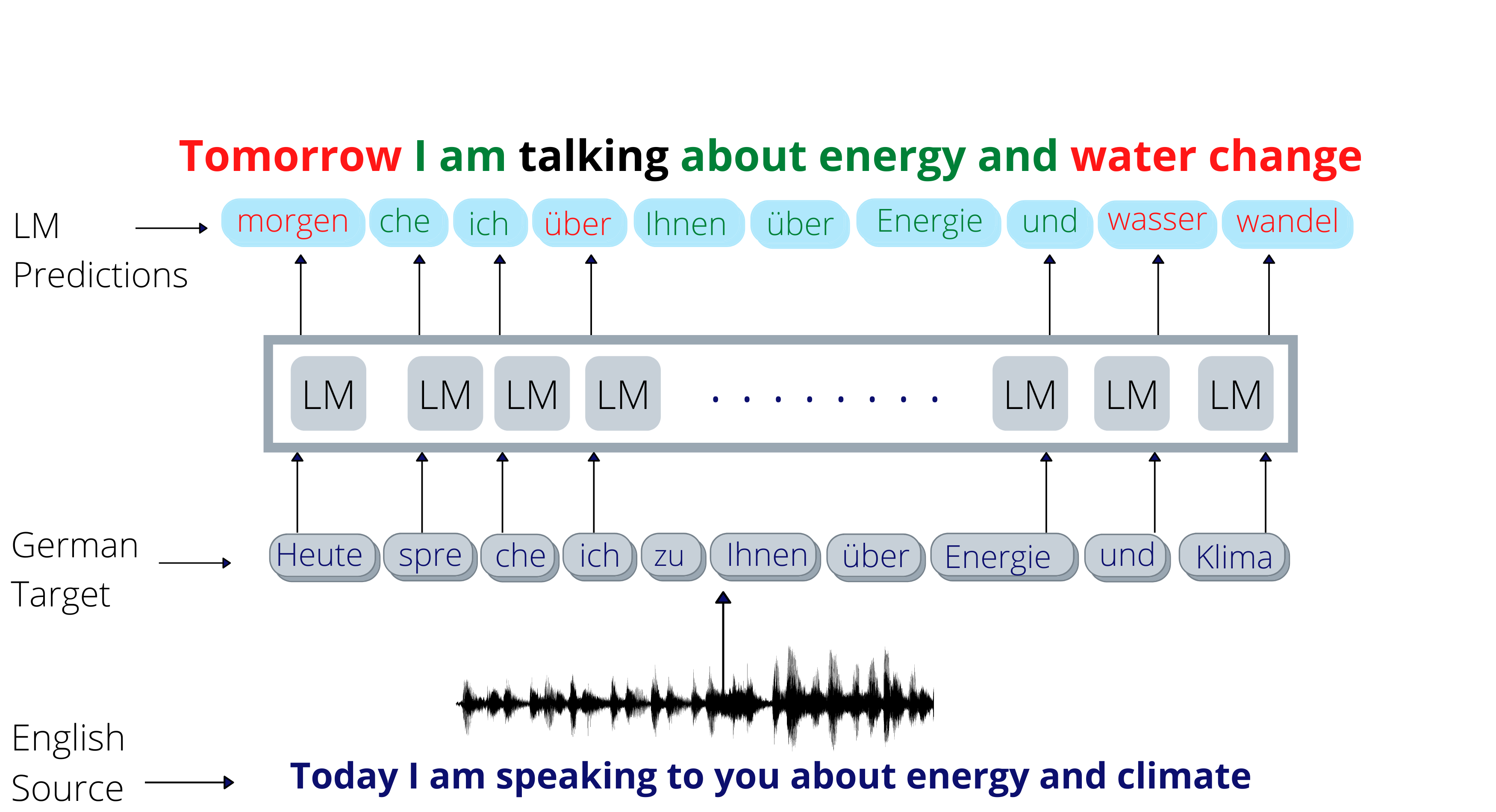}
    \caption{The finetuned XLM-Roberta language model predicts German words using the prefix as input.\\ ({\textcolor{ForestGreen}{Green}}: Correct, {\color{red}Red}: Incorrect, {\color{black}Black}: Neutral).}
    \label{fig:lmex}
\end{figure}

\begin{figure*}[t]
    \centering
    \includegraphics[width=\textwidth]{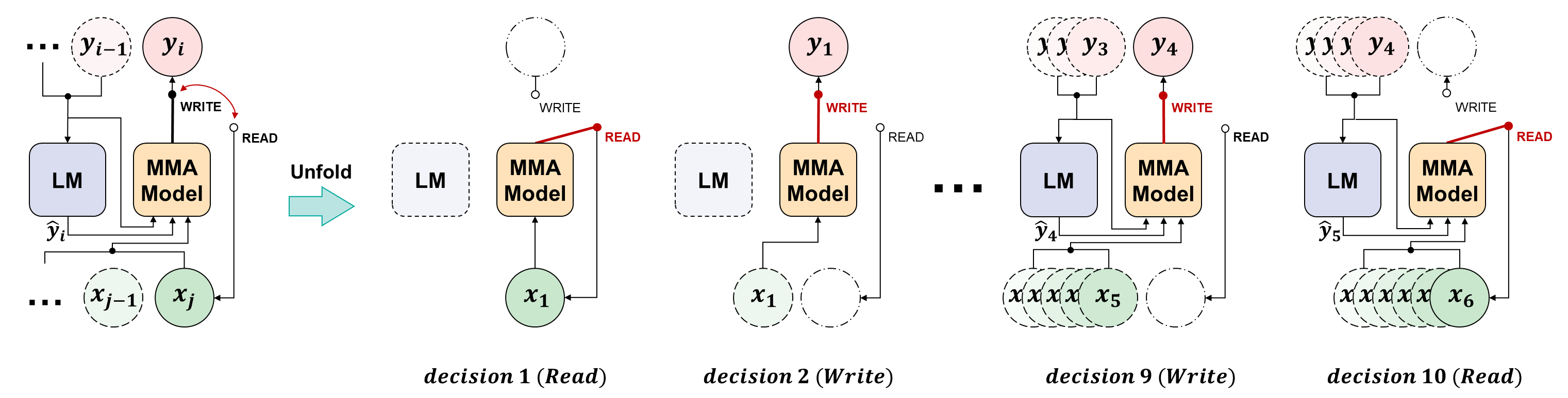}
    \caption{Overview of the simultaneous translation model with future information.}
    \label{fig:mmalmarch}
\end{figure*}

Motivated by human translation experts, we aim to augment monotonic attention with linguistic and extra-linguistic information. We propose to use language model (LM) to obtain the above-mentioned future information. As shown in Figure \ref{fig:lmex}, at each step, the LM takes the prefix target (and source, for cross-lingual LM) sequence and predicts the plausible future information. We hypothesize that aiding the monotonic attention with this future information might help the SNMT model improve the latency-quality trade-off.

The main contributions of this paper are: (1) A novel monotonic attention mechanism to leverage the future information (2) Improved latency-quality trade-offs compared to the state-of-the-art MMA models on MUST-C \cite{di-gangi-etal-2019-must} English-German and English-French speech-to-text translation tasks. (3) Analyses on how our proposed monotonic attention achieves superior performance over MMA and LM rescoring-based MMA. We also analyze the performance of the proposed framework with custom and general-purpose LM.


\section{Monotonic Attention with Future Information Model}
\begin{figure*}[t]
    \centering
    \includegraphics[width=\textwidth]{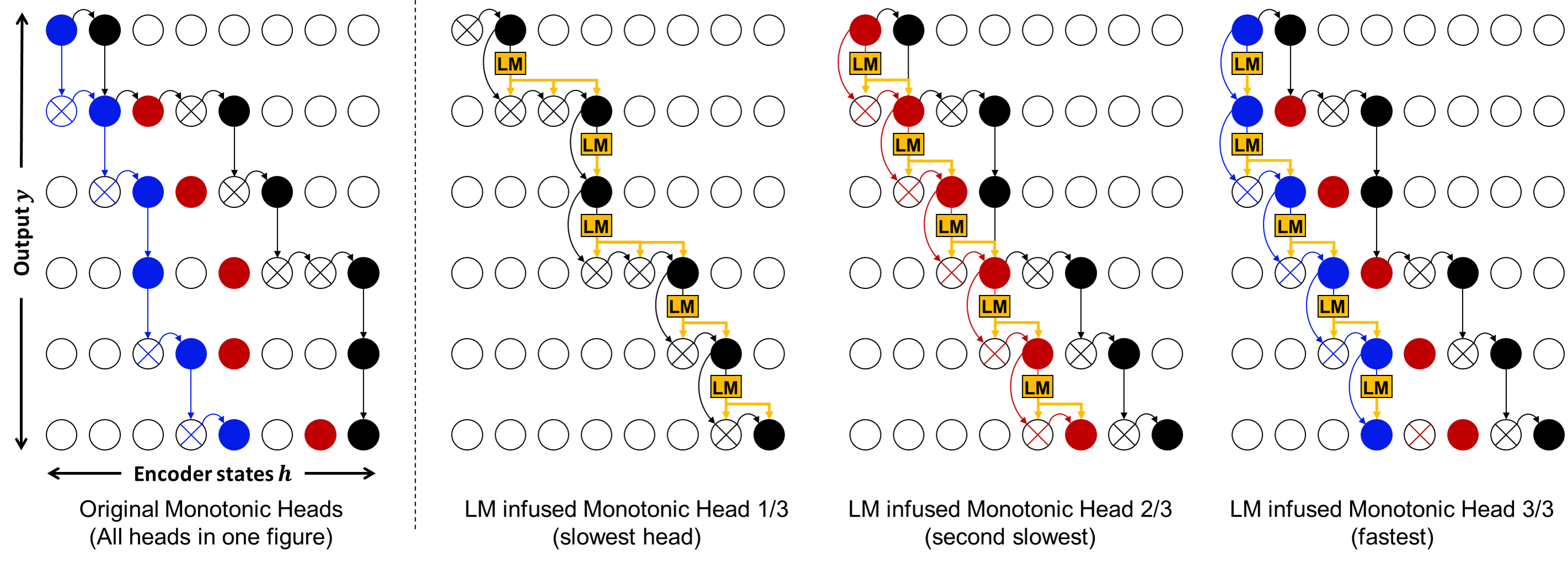}
    \caption{Overview of the monotonic multihead attention with future information.}
    \label{fig:lm-multihead}
\end{figure*}

\begin{algorithm}[t]
\SetAlgoLined
\caption{Monotonic Attention with Future Information}
\label{alg:mafia}
\textbf{Input}: Training examples,  $E = \{x^n, y^n\}_{n=1}^{N}$, hyperparameters such as learning rate ($\alpha$), $\lambda$ (latency), language model (LM) 

Extract plausible future information using LM, $ \hat{y}_{i}^{n} = LM({y_{1}^{n}, \cdots, y_{i-1}^{n}}) \quad \forall i, n$

Append  each $\hat{y}^{n}$ information to target sequence $y^{n}$. New target sequence $\bar{y}^n=\{(y_1^n, \hat{y}_1^n), (y_2^n, \hat{y}_2^n), \cdots, (y_{|y|}^n, \emptyset)\}$

Modified training examples are $\bar{E} = \{x^n, \bar{y}^n\}_{n=1}^{N}$

Initialize model parameters $\theta$

\While{training is not done}{
  Sample a  training example(s) from $\bar{E}$
  
 compute $\{h_{1}, \cdots  
 , h_{|x|}\}$ for $x$ (Eq. \ref{eq:xr})
  
 tokenize $y$ and $\hat{y}_{i}$
 
 Run a sub-token summary layer ( Eq. \ref{eq:tsl}) on   $\hat{y})$ , to obtain $\tilde{\mathbf{y}_i}$ for each $\mathbf{y}_{i-1}$

  \For{each decoding step}{
      Compute monotonic energy and future monotonic energy for $\mathbf{y}_{i-1}$, $\tilde{\mathbf{y}_i}$ (Eq. \ref{eq:memh} and \ref{eq:memhf})
      
      Compute \textit{read/write} decision (Eq. \ref{eq:mrwdf})
      
      compute context vector $c_i$ (Eq. \ref{eq:mmac}), and output target token (Eq. \ref{eq:yr}, \ref{eq:output})
  }
  Compute the latency loss along with negative log likelihood 
  
  Update $\theta$ with gradient descent
  
 }
 
 Return: $\theta$
 
\end{algorithm}

\subsection{Monotonic Attention} 
\label{sec:monotonicattention}
The source and the target sequences are represented as $\mathbf{x}=\{x_1, x_2, \cdots, x_S\}$ and $\mathbf{y}=\{y_1, y_2, \cdots, y_T\}$, with $S$ and $T$ being the length of the source and the target sequences.
The simultaneous machine translation models (SNMT) produce the target sequence concurrently with the growing source sequence. In other words, the probability of predicting the target token $y_i \in \mathbf{y}$  depends on the partial source and target sequences ($x_{\leq j}  \in \mathbf{x}, y_{< i} \in \mathbf{y}$). In this work, we consider sequence-to-sequence based SNMT model in which each target token $y_{i}$  is generated as follows:
\begin{gather}
    h_j = \mathcal{E}(x_{\leq j}) \label{eq:xr} \\
    s_i = \mathcal{D}(y_{<i}, c_i) \label{eq:yr} \\
    y_i = Output(s_i) \label{eq:output}
\end{gather}

\noindent where $\mathcal{E}(.)$ and $\mathcal{D}(.)$ are the Transformer \cite{Transformer} encoder and decoder layers, and $c_i$ is a context vector. In offline MT, the context vector is computed using a soft attention mechanism \cite{Bahdanau14}. In monotonic attention based SNMT, the context vector is computed as follows:
\begin{gather}
    e_{i,j} = MonotonicEnergy(y_{i-1}, h_{j}) \label{eq:me} \\
    p_{i,j} = Sigmoid(e_{i,j}) \label{eq:rwd} \\
    z_{i,j} \sim Bernoulli(p_{i,j}) \label{eq:pd}
\end{gather}


When generating a new target token $y_i$, the decoder chooses whether to \textit{read/write} based on Bernoulli selection probability $p_{i,j}$. When $z_{i,j}=1$, we set $t_i=j$, $c_i=h_j$ (\textit{write}) and generate the target token $y_i$; otherwise, we set $t_i = j+1$ (\textit{read}) and repeat Eq. \ref{eq:me} to \ref{eq:pd}. Here $t_i$ refers to the index of the encoder entry needed to produce the $i^{th}$ target token. Instead of hard assignment of $c_i=h_j$, \citet{raffel17a} compute an expected alignment $\boldsymbol{\alpha}$ which can be calculated in a recurrent manner as shown in Eq. \ref{eq:cumalpha}:

\begin{align}
    \alpha_{i,j} &= p_{i,j} \sum_{k=1}^{j} \left(\alpha_{i-1,k}\prod_{l=k}^{j-1}(1-p_{i,l})\right) \nonumber \\ 
                &= p_{i,j}\left((1-p_{i,j-1})\frac{\alpha_{i,j-1}}{p_{i,j}} + \alpha_{i-1,j}\right) 
    \label{eq:cumalpha}
\end{align}

\noindent \citet{raffel17a} also propose a closed-form parallel solution that allows $\alpha_{i,j}$ to be computed for all $j$ in parallel using cumulative sum and cumulative product operations. \citet{arivazhagan-etal-2019-monotonic} propose Monotonic Infinite Lookback Attention (MILk), which combines the soft attention with monotonic attention to attend all the encoder states from the beginning of the source sequence till $t_i$ for each $y_i$.
\citet{ma2019monotonic} extends MILk to monotonic multihead attention (MMA) to integrate it into the Transformer model \cite{Transformer}. 

The MMA model implements the monotonic energy function in Eq. \ref{eq:me} through scaled-dot product attention. For a Transformer model with $L$ decoder layers and $H$ attention heads per layer,  the  energy function of a $h$-th head encoder-decoder attention in the $l$-th decoder layer is computed as follows:
\begin{gather}
    e_{i,j}^{l,h} = \left( \frac{h_jW_{l,h}^{K}(\mathbf{y}_{i-1}W_{l,h}^Q)^{T}}{\sqrt{d_k}} \right)_{i,j} \label{eq:memh} \\
     p_{i,j}^{l, h} = Sigmoid(e_{i,j}^{l, h}) \label{eq:mhrwd} 
\end{gather} 
\noindent where $W_{l,h}^{K}$ and $W_{l,h}^{Q}$ are the projection matrices for $h_j$ and $\mathbf{y}_{i-1}$, $\mathbf{y}_{i-1}$ is the representation of the previous output token and $d_k$ is the dimension of the attention head. 

The MMA attention for each head is calculated as follows:

\begin{gather}
    u_{i,o}^{l, h} =  \left( \frac{h_oW_{l,h}^{K}(y_{i-1}W_{l,h}^Q)^{T}}{\sqrt{d_k}} \right)_{i,j}, o \in 1, 2, \cdots, t_i \label{eq:mmau} \\
    \beta_{i,j} = \sum_{k=j}^{|x|} \left( \frac {\alpha_{i,k} exp(u_{i,j})}{\sum_{n=1}^{k}exp(u_{i, n})}\right)   \\
    c_i = \sum_{j=1}^{|x|}\beta_{i,j}h_{j} \label{eq:mmac}
\end{gather}

The attention mechanisms in MILk and MMA encourage the model to output the target token with limited source information by adding latency loss metrics to the training objective. Please refer to \citet{arivazhagan-etal-2019-monotonic} and \citet{ma2019monotonic} for more details.

\subsection{Monotonic Attention with Future Information}
The monotonic attention described in Section \ref{sec:monotonicattention}  performs anticipation based only on the currently available source  and target information. We propose to use linguistic and extra-linguistic information, similar to human interpreters, to improve the performance of the SNMT model.  

In order to get the future information for monotonic attention, we rely on LMs, which can inherently provide linguistic information. The extra-linguistic information is also obtained when the LM is finetuned on a particular task. To incorporate the future information, we propose the following modifications to the monotonic attention.

\subsubsection{Future Representation  Layer}
\label{sec:futurerepresentationlayer}
At every decoding step $i$, the previous target token $y_{i-1}$ is equipped with a plausible future token $\hat{y}_{i}$ as shown in the Figure \ref{fig:mmalmarch}. Since the token $\hat{y}_{i}$ comes from an LM possibly with a different tokenizer and vocabulary set, applying the model's tokenizer and vocabulary might split the token $\hat{y}_{i}$ further into multiple sub-tokens $\{\hat{y}_{i}^{1},  \hat{y}_{i}^{2}, \cdots,   \hat{y}_{i}^{m}\}$. To get a single future token representation $\tilde{\textbf{y}}_{i} \in \mathcal{R}^d$ from all the sub-tokens, we apply a  sub-token summary layer as follows:

\begin{equation}
\label{eq:tsl}
\tilde{\textbf{y}}_{i} = \Gamma(\{\hat{y}_{i}^{1},  \hat{y}_{i}^{2}, \cdots,   \hat{y}_{i}^{m}\}) 
\end{equation}

\noindent The $\Gamma$ represents a general sequence representation layer such as a Transformer encoder layer or a simple normalized sum of sub-token representations. 

We enrich $\tilde{\textbf{y}}_{i}$ at every layer $l$ of the decoder block by applying a residual-feed forward network similar to the final sub-layer in the transformer decoder block. 

\begin{equation}
    \label{eq:eftkn}
     \tilde{\mathbf{y}}_{i}^{l} = FFN(\tilde{\mathbf{y}}_{i}^{l-1})
\end{equation}

\subsubsection{Monotonic Energy Layer with Future Information}
We can add the plausible future information to the output layer or append it to the target token representation $y_{i-1}$. However, the MMA \textit{read/write}  decisions happen in Eq. \ref{eq:memh}, therefore, we integrate $\tilde{\mathbf{y}}_{i}$ into the Eq. \ref{eq:memh}.
This way of integration of plausible future information allows the model to condition the LM output usage on the input speech. Hence, it can choose to discard incorrect information. 
The integration is carried out by modifying Eq. \ref{eq:me} - Eq. \ref{eq:mhrwd} in the following way:

First, we compute the monotonic energy for future information using  the enriched future token representation $\tilde{\mathbf{y}}_{i}^{l}$ available at each layer:
\begin{equation}
    \label{eq:memhf}
    \tilde{e}_{i,j}^{l} = \left( \frac{h_j\tilde{W}_{l}^{K}(\tilde{\mathbf{y}}_{i}\tilde{W}_{l}^Q)^{T}}{\sqrt{d}} \right)_{i,j}
\end{equation}
\noindent where $\tilde{W}_{l}^{K}$ and $\tilde{W}_{l}^{Q}$ are the projection matrices for $h_j$ and $\tilde{\mathbf{y}}_{i}$, and $d_k$ is the dimension of the attention head of Eq. \ref{eq:memh}.

We integrate the future monotonic energy function into Eq. \ref{eq:mhrwd} as follows:

\begin{equation}
    \label{eq:mrwdf}
    \tilde{p}_{i,j}^{l, h} = \Omega(e_{i,j}^{l, h}, \tilde{e}_{i,j}^{l}) 
\end{equation}

\noindent $\Omega$ represents a general modulation operator and can be replaced with feature-wise linear modulation \cite{FiLM}  or multiplicative or additive operations. As shown in Figure \ref{fig:lm-multihead}, during training, each head in the monotonic energy  $e_{i,j}^{l, h}$ sees the same future monotonic energy $\tilde{e}_{i,j}^{l}$, since during the inference of multihead monotonic attention, the \textit{write} operation depends on the slowest head.  

After computing $\tilde{p}_{i,j}^{l, h}$, we compute $c_i$ using Eq. \ref{eq:mmac}. The overall process is described in  Algorithm \ref{alg:mafia}. In our experiments, we use the MMA-Infinite Lookback attention model, but our algorithm can be easily extended for MMA-H by modifying the context vector computation to choose only one encoder state.

\subsubsection{Inference}

During the inference time, the \textit{start} token does not contain any plausible information. After predicting the first target token, for every subsequent prediction of target token $y_i$, we invoke the LM to predict the next plausible future token. Whenever new source information arrives, we run step 8 of Algorithm \ref{alg:mafia}. Similarly, for every new target token prediction, we extract new plausible future information from LM using the available predicted sequence, and then we run step 9 to 14 of the Algorithm \ref{alg:mafia}.

\section{Experiments}
    




\subsection{Datasets and Metrics}

We conduct our experiments on the English(En)-German(De) and English(En)-French(Fr) speech-to-text (ST) translation task. These tasks are more involved compared to the text-to-text translations since these are low resource tasks. Moreover, due to different input-output modalities, the difference in source and target sequence lengths is larger compared to the text-to-text task. We use the EnDe, and EnFr portions of the MuST-C dataset. We use the MuST-C dev set for validation and tst-COMMON test set for evaluation. More details about the datasets have been provided in the Appendix.

The speech sequence is represented using 80-dimensional log mel features extracted using the Kaldi\cite{povey2011kaldi} toolkit with a 25ms window size and 10ms shift. The target sequence is represented as subwords using a SentencePiece \cite{kudo2018sentencepiece} model with a unigram vocabulary of size 10,000. We evaluate the performance of the models on both the latency and quality aspects. We use Average Lagging(AL) as our latency metric and case-sensitive detokenized SacreBLEU \cite{post2018call} to measure the translation quality, similar to \cite{ma2020simulmt}. 
The best models are chosen based on the dev set results and reported results are computed using the MuST-C test sets.

\begin{figure*}[t]
    \begin{minipage}[b]{0.49\linewidth}
       \centering
        \centerline{\includegraphics[height=5.8cm]{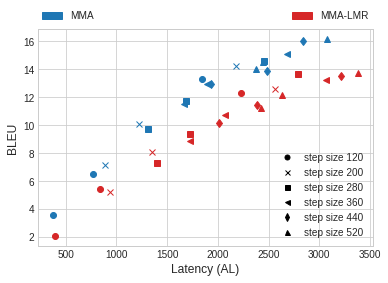}}
        \centerline{(a) EnDe Task }\medskip
        \label{fig:de_results}
    \end{minipage}
    \begin{minipage}[b]{0.49\linewidth}
       \centering
        \centerline{\includegraphics[height=5.8cm]{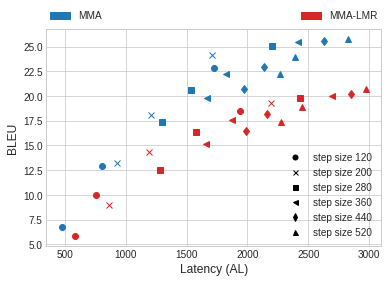}}
        \centerline{(b) EnFr Task }\medskip
        \label{fig:fr_results}
    \end{minipage}
        \caption{\label{fig:lm_rescore}
    BLEU vs Average Lagging results for MMA and MMA-LMR models}
\end{figure*}


\subsection{Implementation Details}
Our base model is adopted from \citet{ma2020simulmt} and the initial implementation is taken from Fairseq\footnote{\href{https://github.com/pytorch/fairseq}{https://github.com/pytorch/fairseq}} repository. In the text-to-text case, each encoder state corresponds to a vocabulary unit. Hence, the \textit{read/write} decisions are taken for each encoder state. For Simultaneous ST, each encoder state represents only 40ms of speech, assuming a sub-sampling factor of 4 from the convolutional layers. We use a pre-decision ratio of 7 (segment size of 280ms), which means that the simultaneous \textit{read/write} decisions are made after every seven encoder states, which roughly corresponds to a word(average word length for the EnDe dataset is 270ms \cite{ma2020simulmt}). Since we train MMA-IL \cite{ma2019monotonic} models, we set $\lambda_{var}=0$ for all our experiments as it was not reported to be helpful for models with infinite lookback. We use $\lambda$ or $\lambda_{latency}$ to refer to the hyperparameter corresponding to the weighted average($\lambda_{avg}$) in MMA. The values of this hyperparameter $\lambda$ are chosen from the set $\{0.01,0.05,0.1\}$. The $\Gamma$ layer in Eq. \ref{eq:tsl} computes the normalized sum of the sub-token representations. For SLM, it simply finds the embedding since it shares the same vocabulary set. The $\Omega$ layer in Eq. \ref{eq:mrwdf} performs the additive operation to add the energies corresponding to previous output token $y_{i-1}$ and the prediction $\hat{y}_i$. All the models were trained by simulating 8 GPU settings on a single NVIDIA v100 GPU. 

\subsection{Models} We train a baseline model based on \citet{ma2020simulmt}, called the MMA  model. The base MMA model encoder and decoder embedding dimensions are set to 392, whereas our proposed model's encoder and decoder embeddings are set to 256 to have similar parameters ($ \approx39M$)  for a fairer comparison.  Apart from the encoder and decoder embedding dimension difference, all other hyperparameter settings and training procedures are similar for all the reported models. We train two MMA models based on two different LMs used for extracting future information. Details have been provided in Section \ref{sec:lm}.

We explored a naive approach of integrating LM information into the MMA. We modify the baseline MMA  inference process to integrate the information from the LM.  Instead of greedy decoding, we evaluate the top-5 outputs from the MMA at each decoding step using the LM. We use the perplexity score provided by the LM to rank the partial predicted sentences formed using these top-5 outputs and choose the best one. The advantage of this approach is that we do not need to modify the training process of the MMA model. We refer to this experiment as 'LM Rescoring(LMR)', and the corresponding model is called MMA-LMR.

We follow the training process similar to \citet{ma2020simulmt} training process.  We train an English ASR model using the source speech data. Next, we train a simultaneous model without the latency loss (setting $\lambda_{latency}=0$) after initializing the encoder from the English ASR model. After this step, we finetune the simultaneous model for different $\lambda$s. This training process is repeated for all the reported models and for each task. 

\begin{figure}[t]
     \centering
     \includegraphics[width=0.5\textwidth]{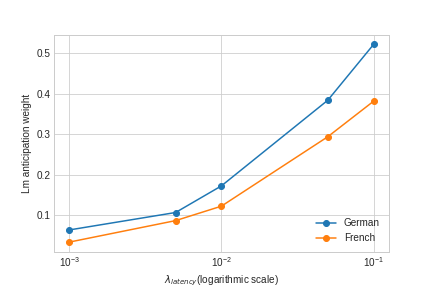}
     \caption{LM prediction weight vs $\lambda$}
     \label{fig:lm_weight}
 \end{figure}
 
\subsection{Language Model}
\label{sec:lm}
We use two different language models to train our proposed LM-based simultaneous speech translation model. Firstly, we use the pretrained XLM-RoBERTa \cite{conneau2019unsupervised} model from Huggingface Transformers\footnote{\href{https://huggingface.co/transformers/}{https://huggingface.co/transformers/}} model repository. It is a multilingual(more than 100 languages) LM trained on a wide variety of cross-lingual tasks. The model contains 550M parameters with 24 layers, 1,024 hidden-states. Since the LM output can be very open-ended and might not directly suit/cater to our task and dataset, we finetune the head of the model using the target MuST-C text data corresponding to each task.

We also train a smaller language model(SLM), which contains 6 Transformer decoder layers, 512 hidden-states and 24M parameters. We use the MuST-C data along with additional data augmentation to reduce overfitting. Although trained on a much smaller monolingual dataset, it helps to remove the issues related to vocabulary mismatch as discussed in the Section \ref{sec:futurerepresentationlayer}. This LM has lower inference time and has higher accuracy on the next token prediction task as compared to XLM, 30.15\% vs. 21.5\% for German \& 31.65\% vs. 18.45\% for French. More details about LMs have been provided in the Appendix. The models trained using these LMs are referred to as MMA-XLM and MMA-SLM.


\subsection{Results}
In this section, we first provide the results for the MMA-LMR model. The results for MMA-XLM and MMA-SLM have been provided in the form of latency-quality trade-off curves. Finally, we provide analysis and explanation for the results.

\paragraph{LM based Rescoring}
For MMA-LMR, we use the LM only during inference. As observed in Figure \ref{fig:lm_rescore}, MMA-LMR has inferior performance compared to the MMA model. Since  the LM information integration is not conditioned on the input speech, the MMA model cannot discard the incorrect information from LM. This motivates us to tightly integrate the LM information into the simultaneous model.

\begin{figure*}[t]
    \begin{minipage}[b]{0.49\linewidth}
       \centering
        \centerline{\includegraphics[height=5.8cm]{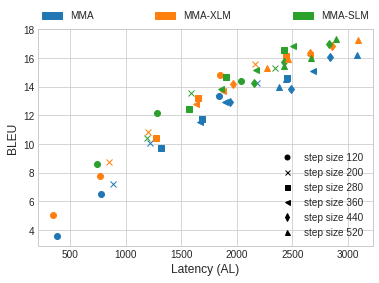}}
        \centerline{(a) EnDe Task }\medskip
        \label{fig:de_results}
    \end{minipage}
    \begin{minipage}[b]{0.49\linewidth}
       \centering
        \centerline{\includegraphics[height=5.8cm]{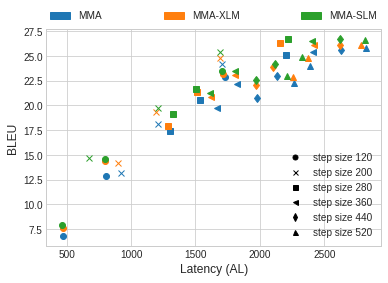}}
        \centerline{(b) EnFr Task }\medskip
        \label{fig:fr_results}
    \end{minipage}
        \caption{\label{fig:results}
    BLEU vs Average Lagging results for MMA, MMA-XLM and MMA-SLM models. MMA-XLM and MMA-SLM provide better latency/quality tradeoff curves.}
\end{figure*}

\paragraph{Latency-Quality Curves}
We obtain several simultaneous models based on MMA, MMA-XLM, MMA-SLM systems  by using different latencies ($\lambda$) during training and different speech segment sizes during inference. The speech segment size refers to the duration of speech(in ms) processed corresponding to each \textit{read} decision. We plot the BLEU scores against the Average Lagging(AL) incurred during the evaluation using SimulEval toolkit\cite{simeval}. The BLEU-AL curves for all the models have been provided in Figure \ref{fig:results}. We can observe that the LM-based models provide a better translation quality at the same latency or lower latency for the same translation quality or some times improving in both aspects

\paragraph{LM anticipation vs Latency}
In order to measure the relative weight given to the predictions from the LM, we compare the norm of the monotonic energies corresponding to the LM predictions $e_{pred}$(Eq. \ref{eq:memhf}) and the previous output tokens $e_{output}$(Eq. \ref{eq:memh} ). Let us define LM prediction weight as 
\newcommand\norm[1]{\left\lVert#1\right\rVert}

\begin{equation}
    \label{eq:lmpw}
    LM_{pw} = \left( \frac{\norm{e_{pred}}}{\norm{e_{output}}} \right)
\end{equation}
In Figure \ref{fig:lm_weight}, we plot the variation of $LM_{pw}$(averaged) vs. ${\lambda}$. We can observe that as the latency requirements become more and more strict, the model starts to give more weightage to the predictions coming from the LM. In other words, as the need for anticipation increases, the model starts to rely more on the LM predictions. 

\paragraph{Effect of LM Size on CAAL}
As we can observe from the Figure \ref{fig:results}, the results for MMA-XLM and MMA-SLM are very close. Not only does MMA-SLM perform slightly better than MMA-XLM, but the prediction computation time is also much lower since XLM is much deeper than SLM. The average time taken to compute one token during prediction for SLM is 6.1ms as compared to 24.78ms for XLM. However, XLM is more suitable for multilingual settings since we can use same LM for all the languages.

\begin{figure*}[t]
    \begin{minipage}[b]{0.4\textwidth}
       \centering
        \centerline{\includegraphics[height=5.8cm]{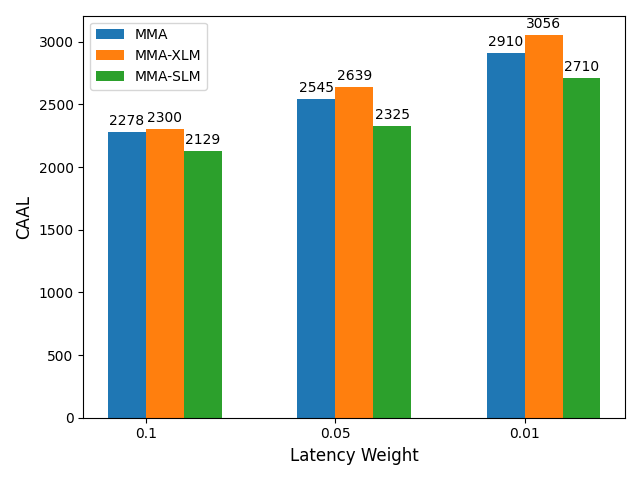}}
        \centerline{(a) EnDe Task}\medskip
        \label{fig:alca_de}
    \end{minipage}
    \hspace{2.0cm}
    \begin{minipage}[b]{0.4\textwidth}
       \centering
        \centerline{\includegraphics[height=5.8cm]{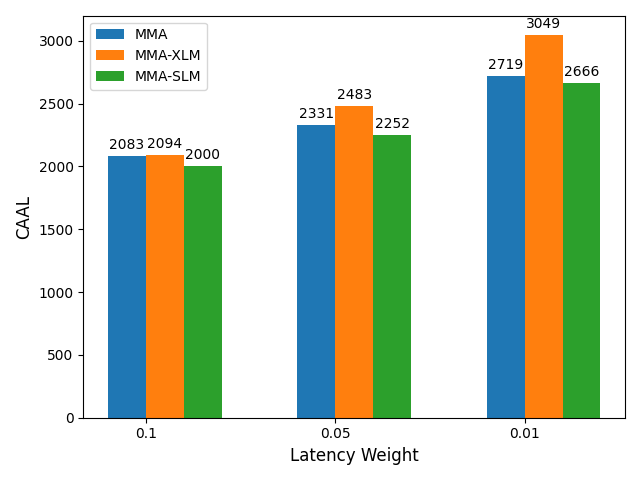}}
        \centerline{(b) EnFr Task }\medskip
        \label{fig:alca_fr}
    \end{minipage}
   \caption{\label{fig:alca}
    Computational aware latency of MMA, MMA-XLM, MMA-SLM with similar BLEU scores and different latencies=\{0.1, 0.05, 0.01\}}
\end{figure*}

In order to account for the computation time incurred by the model, we also use the Computation Aware Average Latency (CAAL) introduced in \cite{ma2020simulmt}. AL(non-computation aware) is measured in terms of the duration of speech listened to before generating target token, while CAAL uses the wall-clock time, which also factors in the time elapsed due to the model complexity. 
In Figure \ref{fig:alca}, we provide the CAAL for various $\lambda$ values for approximately similar BLEU. For a given value of AL, LM based MMA models have a higher CAAL when compared to the MMA model. This gap is expected and occurs due to the time taken for the LM to compute the predictions. However, both MMA-XLM and MMA-SLM improve the latency-quality trade-off and hence reduce the AL for a given BLEU. As observed, MMA-SLM has lesser CAAL as compared to MMA since the extra computation time is balanced by the reductions in AL due to the algorithmic improvements. MMA-XLM, on the other hand, has a slightly higher CAAL. 
English speakers utter 6.2 syllables per sec\cite{pellegrino2011cross}, which means 160ms/syllable. Both German and French readers read roughly 5 syllables/sec, 200ms/syllable \cite{trauzettel2012standardized}. Considering this human perception speed, a gap of 6 or 24ms per sub-word(which might contain multiple syllables) should not cause acute deterioration in the user experience. 

\section{Related Work}
The earlier works in streaming simultaneous translation  such  as  \citet{cho2016can,gu2016learning,press2018you} lack the ability to anticipate the words with missing source context. \citet{ma2019stacl} established a more sophisticated approach by integrating their \textit{read/write} agent directly into MT. Similar to \citet{dalvi-etal-2018-incremental}, they employ a fixed agent that first reads $k$ source tokens and then proceeds to alternate between \textit{write} and \textit{read} until the source tokens are finished. 
Recently, the adaptive policies based on several variants of monotonic attention for SNMT have been explored: hard monotonic attention \cite{raffel17a}, monotonic chunkwise attention (MoChA) \cite{chiu*2018monotonic} and monotonic infinite lookback attention (MILk) \cite{arivazhagan-etal-2019-monotonic}. MILK improves upon the wait-k training with an attention that can adapt how it will wait based on the current context. 
Monotonic multihead attention (MMA) \cite{ma2019monotonic}  extends MILk to transformer-based models. 

\citet{gulcehre2015using, 9a7e5bdf2b9b4df8b46c996f56c5b884} propose shallow fusion of language models(LMs) into text-to-text machine translation(MT) by combining LM and MT scores at inference time using a  log-linear model. However, this has a mismatch between training and inference of MT. To overcome this drawback, \cite{stahlberg-etal-2018-simple} integrate the LM scores during MT training. They use a pretrained LM and train the MT system to optimize the combined score of LM and MT on the training set. \cite{Sriram2018} explored a similar idea for Automatic Speech Recognition(ASR) using a gating network for controlling the relative contribution of the LM. These techniques allow the main sequence to sequence model(either ASR or MT) to focus on modeling the source sentence, while the LM controls the target side generation. 
\citet{wu2020learn} implicitly uses future information during training of SNMT systems by simultaneously training different wait-k systems. The translation model is jointly trained with a controller model that decides which  $k$ is optimal to use for training a particular batch of examples. However, they do not use any explicit future information during training and inference. 

End-to-end speech translation \cite{our_icassp20, sperber2020speech} has recently made great progress and even surpassed the cascaded models (ASR followed by MT) \cite{ney1999speech,cascade1}.
Recently, \citet{simulspeech} investigate how to adapt the simultaneous models for speech-to-text translation tasks. \citet{han-etal-2020-end} use meta-learning algorithm \cite{pmlr-v70-finn17a} and improve these wait-k based simultaneous speech-to-text models. \citet{ma2020simulmt} explore the usage of monotonic multihead attention by introducing a pre-decision module.


\section{Conclusion}
In this work, we provide a generic framework to integrate the linguistic and extra-linguistic information into simultaneous models. This information helps to improve the anticipation for monotonic attention based SNMT models. We rely on language models to extract plausible future information and propose a new monotonic attention mechanism to infuse future information. We conduct several experiments on low resource speech-to-text translation tasks to show the effectiveness of proposed approach. We have achieved superior quality-latency trade-off compared to the state-of-the-art monotonic multihead attention. In the future work, we plan to extend the proposed framework to the text-to-text simultaneous translation and analyze different future information fusion mechanisms.      

\bibliography{anthology,custom}
\bibliographystyle{acl_natbib}

\newpage

\appendix
\begin{table*}[t]
\centering
\begin{tabular}{ |c|c|c|c|c|c|c|c|}
 \hline
 \multirow{2}{*}{Task} & \multirow{2}{*}{\# Hours}  & \multicolumn{3}{c|}{\# Sentences} &  \multirow{2}{*}{\# Talks}  & \multicolumn{2}{c|}{\# Words}  \\  \cline{3-5} \cline{7-8}
 
 \multirow{2}{*}{} & \multirow{2}{*}{}  & Train & Dev & Test & \multirow{2}{*}{} & Source & Target    \\ 
 \hline \hline
 English-German & 408 & 225k & 1,423 & 2,641 & 2,093 & 4.3M & 4M \\
 English-French & 492 & 269k & 1,412 & 2,632 & 2,510 & 5.2M & 5.4M \\
 \hline

\end{tabular}

\caption{Dataset Statistics(\small{\# - Number of})}
\label{table:data}
\end{table*}

\begin{table*}[!]
\centering
\begin{tabular}{ c c c}
 \hline
                & \multirow{2}{*}{MMA} & \multirow{2}{*}{MMA-XLM/CLM} \\
 Hyperparameter &  &  \\ \hline \hline
 encoder layers & 12  & 12\\
 encoder embed dim & 292 & 256\\
 encoder ffn embed dim & 2048 & 2048 \\
 encoder attention heads & 4 & 4\\
 decoder layers & 6 & 6 \\
 decoder embed dim & 292 & 256 \\
 decoder ffn embed dim & 2048 & 2048 \\
 monotonic ffn embde dim & -- & 2048 \\
 decoder attention heads & 4  & 4\\
 dropout & 0.1 & 0.1 \\
 optimizer & adam & adam \\
 adam-$\beta$ & (0.9, 0.999)  & (0.9, 0.999) \\
 clip-norm & 10.0  & 10.0 \\
 lr scheduler & inverse sqrt  & inverse sqrt \\
 learning rate & 0.0001  & 0.0001 \\
 warmup-updates & 4000  & 4000 \\
 label-smoothing & 0.0 & 0.0 \\
 max tokens & 40000 & 40000 \\
 conv layers & 2  & 2 \\
 conv stride & (2,2) & (2,2) \\
 \#params     & $\approx39M$ & $\approx39M$ \\
 
 \hline
\end{tabular}

\caption{Model Hyperparameters}
\label{table:hparams}
\end{table*}

\section{Language Models}
As mentioned earlier, we train two different language models (LMs) and use them to improve the anticipation in monotonic attention based Simultaneous models.

\subsection{XLM-Roberta(XLM-R)\footnote{\href{https://huggingface.co/xlm-roberta-large}{https://huggingface.co/xlm-roberta-large}}} XLM-R Large model was trained on the 100 languages CommonCrawl corpora total size of 2.5TB with 550M parameters from 24 layers, 1024 hidden states, 4096 feed-forward hidden-states, and 16 heads. Total number of parameters is 558M. We finetune the head of the XLM-R LM model using the Masked Language Modeling objective which accounts for 0.23\% of the total model parameters, i.e., 1.3M parameters.

\subsection{Smaller Language Model}
Since the LM predictions are computed serially during inference, the time taken to compute the LM token serves as a bottleneck to the latency requirements. To reduce the LM computation time, we train a smaller Language Model (SLM) from scratch using the Causal Language Modeling objective. SLM is composed of 6 Transformer decoder blocks, 512 hidden-states, 2048 feed-forward hidden-states \& 8 attention heads. It alleviates the need for the sub-token summary layer since it shares the vocabulary and tokenization with the MMA models. The train examples are at the sentence level, rather than forming a block out of multiple sentences(which is the usual case for Language Models).

Since the target texts contain lesser than 250k examples, we use additional data augmentation techniques to upsample the target data. We also use additional data to avoid overfitting on the MuST-C target text. Details have been provided in \ref{sec:da}.

\subsubsection{Data Augmentation}
\label{sec:da}

\paragraph{Up-Sampling:} To boost the LM performance and mitigate overfitting, we use contextual data augmentation \cite{kobayashi-2018-contextual} to upsample the MuST-C target text data by substituting and inserting words based on LM predictions. We use the NLPAUG \footnote{\href{https://pypi.org/project/nlpaug/}{https://pypi.org/project/nlpaug/}} package to get similar words based on contextual embeddings. From the Hugging Face Repository, we use two different pretrained BERT \cite{devlin-etal-2019-bert} models for German \textit{bert-base-german-dbmdz-cased} \& \textit{bert-base-german-dbmdz-uncased} and \textit{bert-base-fr-cased} for French. We upsample German to 1.13M examples and French to 1.38M examples.

\paragraph{Additional Data:}
We also use additional data to avoid overfitting. For German we use the Newscrawl(WMT 19) data which includes 58M examples. For French, we use Common Crawl and Europarl to augment 4M extra training examples. 

We observe that both upsampling and data augmentation help us to reduce the overfitting on the MuST-C dev set.

\subsection{Token Prediction}
For each output token, the LM prediction is obtained by feeding the prefix upto that token to the LM model. These predictions are pre-computed for training and validation sets. This ensures parallelization and avoids the overhead to run the LM simultaneously during the training process. During inference, the LM model is called every time a new output token is written.


\section{Dataset}
The MuST-C dataset comprises of English TED talks, the translations and transcriptions have been aligned with the speech at sentence level. Dataset statistics have been provided in the Table \ref{table:data}. 

\section{Hyperparameters}
The details regarding the hyperparameters for the model have been provided in Table \ref{table:hparams}.



\end{document}